%% file: main.tex
\documentclass[10pt,journal,compsoc]{IEEEtran}
\newif\ifpeerreview

\peerreviewfalse
\usepackage{pdfpages}

\usepackage[nocompress]{cite}
\usepackage{url}
\usepackage{amsmath,amssymb,graphicx}
\usepackage{booktabs}

\usepackage[switch]{lineno}

\usepackage{xcolor}
\input{preamble}

\title{TranSplat: Instant Object Relighting in Gaussian Splatting via Spherical Harmonic Radiance Transfer}

\author{Boyang (Tony) Yu$^*$, Yanlin Jin$^*$, Yun He, Akshat Dave, Ravi Ramamoorthi, and Guha Balakrishnan%
\IEEEcompsocitemizethanks{
\IEEEcompsocthanksitem $^*$Equal contribution.
\ifpeerreview \else
\IEEEcompsocthanksitem Project page and code release:
{\color{magenta}\protect\url{https://tonyyu0822.github.io/transplat/}}
\fi
\IEEEcompsocthanksitem B.~Yu, Y.~Jin, and G.~Balakrishnan are with Rice University. E-mail: \{by12, yj56, guha\}@rice.edu
\IEEEcompsocthanksitem Y.~He is with the University of Maryland. E-mail: yunhe@umd.edu
\IEEEcompsocthanksitem A.~Dave is with Stony Brook University. E-mail: dave@cs.stonybrook.edu
\IEEEcompsocthanksitem R.~Ramamoorthi is with the University of California, San Diego. E-mail: ravir@ucsd.edu
}%
}

\begin{document}

\IEEEtitleabstractindextext{%
\input{sec/0_abstract}

\begin{IEEEkeywords}
Gaussian Splatting, Relighting, Spherical Harmonics, Radiance Transfer, Computational Photography
\end{IEEEkeywords}
}

% Make Title
\ifpeerreview
\linenumbers \linenumbersep 15pt\relax
\author{Paper ID \paperID\IEEEcompsocitemizethanks{\IEEEcompsocthanksitem This paper is under review for ICCP 2026 and the PAMI special issue on computational photography. Do not distribute.}}
\markboth{Anonymous ICCP 2026 submission ID \paperID}%
{}
\fi
\maketitle

% Teaser figure
% \begin{figure*}[t!]
% \centering
% \includegraphics[width=\linewidth]{imgs/teaser.pdf}
% \caption{\textbf{Example demonstrations of \name{} relighting a LEGO bulldozer from a source environment (top left) to target environments (bottom left).} We assume that 3D Gaussian Splatting (GS)~\cite{kerbl3Dgaussians} representations may be built for both the source and target scenes from multiple views (not shown here), and that the object may be extracted (segmented) and placed into the target scene with either automatic or manual supervision. \name{} then relights the object's Gaussians to reflect appropriate shading and shadow effects for the target environment. Crucially, \name{} requires no ground truth information on the source or target radiance distributions (i.e., environment maps), no decomposition of object materials, and uses lightweight operations taking \textbf{\emph{1 seconds or less}} of processing time.}
% \label{fig:teaser}
% \end{figure*}

\input{sec/1_intro}
\input{sec/2_related}
% \input{sec/background}
\input{sec/3_methods}

\input{sec/4_experiments}
\input{sec/5_discussion}

% Any acknowledgments to only be included in camera ready
\ifpeerreview \else
\section*{Acknowledgments}
This research is based upon work supported by the Office of the Director of National Intelligence (ODNI), Intelligence Advanced Research Projects Activity (IARPA), via IARPA R\&D Contract No. 140D0423C0076. The views and conclusions contained herein are those of the authors and should not be interpreted as necessarily representing the official policies or endorsements, either expressed or implied, of the ODNI, IARPA, or the U.S. Government. The U.S. Government is authorized to reproduce and distribute reprints for Governmental purposes notwithstanding any copyright annotation thereon. We also acknowledge support from ONR grant N00014-23-1-2526 and the Ronald L. Graham Chair at the University of California, San Diego.
\fi

\bibliographystyle{IEEEtran}
\bibliography{main}

\includepdf[pages=-]{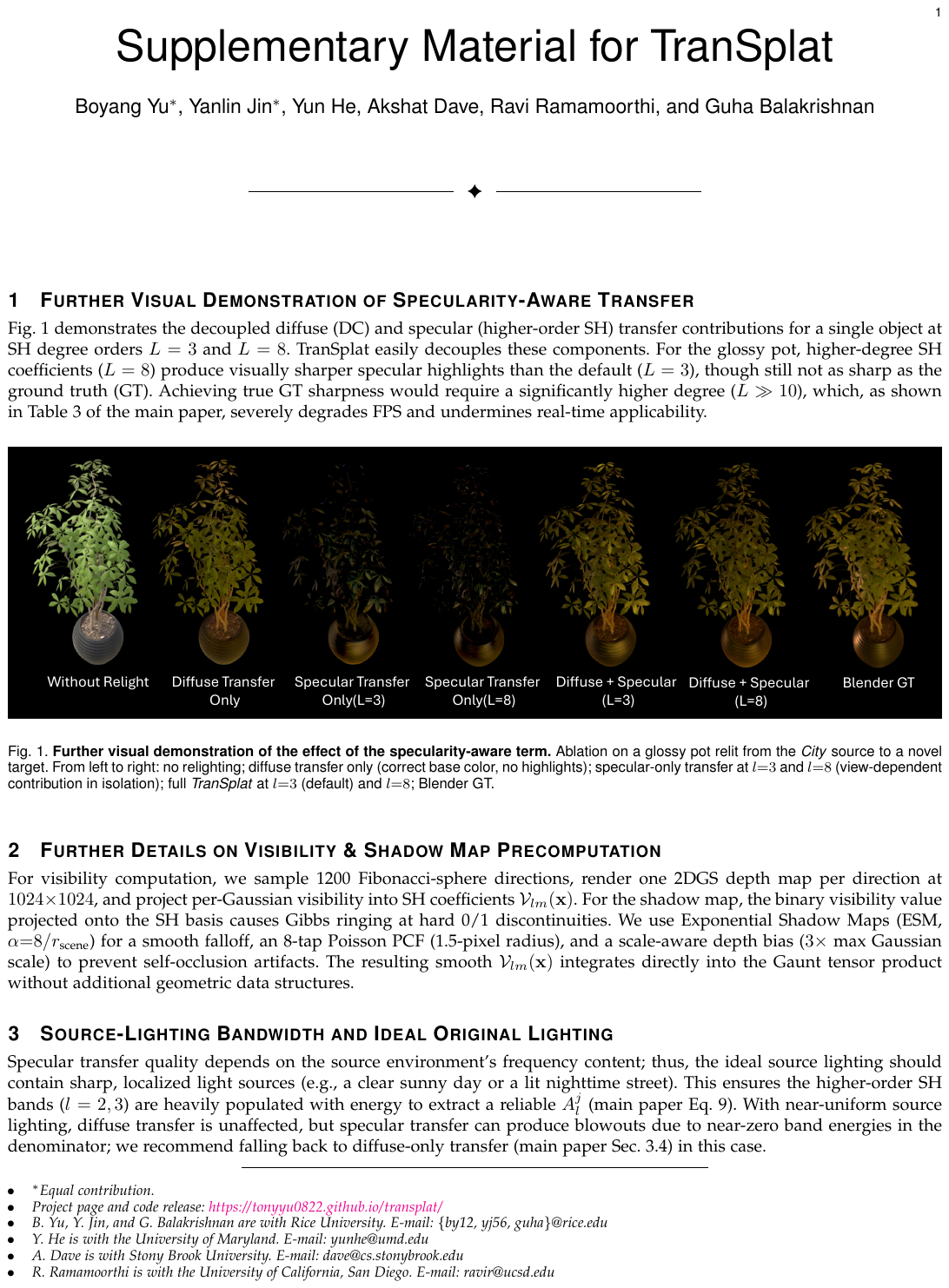}
\end{document}

%% file: preamble.tex
\newcommand{\name}{\emph{TranSplat}}

\usepackage{multirow}
\usepackage[table]{xcolor} % for z\cellcolor
\usepackage{bbm}

\definecolor{BestGreen}{RGB}{180,235,180}   % soft leaf green
\definecolor{SecondOrange}{RGB}{255,245,180} % warm apricot

\newcommand{\best}[1]{\cellcolor{BestGreen}#1}
\newcommand{\second}[1]{\cellcolor{SecondOrange}#1}
\usepackage[table]{xcolor}
\usepackage{booktabs}
\usepackage{multirow}
\usepackage{colortbl}
\usepackage{makecell}

\definecolor{headergray}{RGB}{242,244,247}
\definecolor{groupgray}{RGB}{248,249,250}
\definecolor{oursblue}{RGB}{231,243,255}
\definecolor{bestgreen}{RGB}{234,247,236}

%% file: sec/0_abstract.tex
\begin{abstract}
We present \name{}, a method for instant, accurate object relighting within the Gaussian Splatting (GS) framework. Rather than relying on costly inverse rendering routines, we propose a BRDF-free radiance transfer strategy that analytically modulates the spherical harmonic (SH) appearance coefficients of an object's 2D Gaussian surfels using per-normal irradiance ratios derived from source and target environment maps. To handle view-dependent and glossy appearances without explicit material estimation, we introduce a specularity-aware dual-path SH transfer strategy that adapts higher-order SH bands in the reflection domain. Additionally, we propose a lightweight SH-domain self-shadowing module to ensure physically realistic occlusion without explicit mesh raycasting. Operating as a post-processing step, \name{} requires no additional GS retraining for a pair of source and target scenes. Evaluations on synthetic and real-world objects demonstrate state-of-the-art accuracy, outperforming recent inverse-rendering and diffusion-based GS relighting methods across most conditions, all while completing relighting operations in under one second. Although bounded by radially-symmetric BRDF approximations and the low-pass nature of the SH basis, \name{} produces perceptually realistic renderings even for glossy, complex materials, establishing a valuable, lightweight path forward for GS relighting.

% We present TranSplat, a 3D scene rendering algorithm that enables realistic cross-scene object transfer (from a source to a target scene) based on the Gaussian Splatting framework. Our approach addresses two critical challenges: (1) precise 3D object extraction from the source scene, and (2) faithful relighting of the transferred object in the target scene without explicit material property estimation. TranSplat fits a splatting model to the source scene, using 2D object masks to drive fine-grained 3D segmentation. Following user-guided insertion of the object into the target scene, along with automatic refinement of position and orientation, TranSplat derives per-Gaussian radiance transfer functions via spherical harmonic analysis to adapt the object's appearance to match the target scene's lighting environment. This relighting strategy does not require explicitly estimating physical scene properties such as BRDFs. Evaluated on several synthetic and real-world scenes and objects, TranSplat yields excellent 3D object extractions and relighting performance compared to recent baseline methods and visually convincing cross-scene object transfers. We conclude by discussing the limitations of the approach. 
\end{abstract}

%% file: sec/1_intro.tex
\section{Introduction}
The Gaussian Splatting (GS) framework has demonstrated remarkable power and computational speed for high-quality, real-time neural rendering applications~\cite{kerbl3Dgaussians}. Building on these rendering capabilities, there is a growing demand for realistic 3D scene editing and composition applications with GS, such as object relighting. Given a pretrained Gaussian asset alongside high dynamic range (HDR) environment maps for both the source and target environments, the goal of object relighting is to re-render the object so that its shading and color are consistent with the target illumination settings. The key challenge of this task is that the object's perceived illumination must change as a function of both environments in a physically consistent manner, without being given any knowledge of its material properties.

Several recent GS relighting methods offer solutions to this problem, but rely on computationally heavy paradigms. These include explicit inverse rendering pipelines that attempt to explicitly decompose the scene into geometry, materials (BRDFs), and lighting, which requires slow, iterative per-scene optimization~\cite{jiang2024gaussianshader,liang2023gs,guo2024prtgsprecomputedradiancetransfer}. Alternatively, some methods rely on generative diffusion priors that distill 2D image-space relighting into 3D representations. These often suffer from low resolution, multi-view inconsistencies, and slow inference~\cite{jin2024neural_gaffer}. For example, even the fastest of these optimization-based inverse rendering methods~\cite{liang2023gs,li2025recap} take approximately 60 minutes to jointly optimize and relight a typical scene. This computational burden precludes real-time applications such as gaming, VR, and interactive 3D scene editing.

%expressive scene modeling, they typically rely on computationally heavy paradigms. These include explicit inverse rendering pipelines that require slow, iterative per-scene optimization~\cite{jiang2024gaussianshader,liang2023gs,guo2024prtgsprecomputedradiancetransfer}, or generative diffusion priors that suffer from multi-view inconsistency and slow inference~\cite{jin2024neural_gaffer}.

We propose \name{}, a novel framework that instantly and realistically relights off-the-shelf Gaussian assets in novel target environments, running in under $1$ second on an NVIDIA A6000 GPU. To achieve this, we take inspiration from a theoretical identity relating the spherical harmonic (SH) coefficients of surfaces with radially symmetric BRDFs under different lighting conditions~\cite{Mahajan2006}. For diffuse surfaces, this identity states that relighting reduces to multiplying the SH appearance coefficients by the per-normal irradiance ratios between the target and source environments. Crucially, this \emph{radiance transfer} bypasses explicit BRDF estimation, avoiding a major computational bottleneck. While radiance transfer has been leveraged in classical 2D ratio image methods~\cite{Troccoli2006RecoveringIA, 908964}, it has not been integrated into GS frameworks for lightweight relighting. Our key insight is that the spherical harmonic coefficients of a standard pretrained GS asset, despite entangling source illumination, material response, and view-dependent appearance, can be directly transformed to fit the target environment without retraining or explicit BRDF decomposition.

\name{} introduces three main technical contributions. First, we formulate an analytical diffuse irradiance transfer approach in the GS framework. We enable this by building upon the 2D Gaussian Surfels representation\cite{Huang2DGS2024}, exploiting its high-quality, flat surface normals which are essential for accurate, per-surfel irradiance computation. Second, we extend this formulation beyond the diffuse case. By performing a reflection-domain reparameterization, we extract an approximate BRDF attenuation profile directly from the pretrained SH coefficients, allowing us to modulate glossy appearances without parametric material estimation. Third, we introduce an SH-domain self-shadowing module utilizing smooth continuous visibility functions and the Gaunt tensor, ensuring physically realistic occlusion without explicit raycasting. The entire pipeline operates as a single, closed-form analytical pass over the Gaussian assets. To our knowledge, \name{} is the first framework to enable real-time relighting of off-the-shelf Gaussian assets. While this lightweight method makes certain simplifications at the cost of modeling higher-order, subtle material details, it achieves fast, perceptually realistic relighting of both diffuse and glossy objects directly in the SH domain.

We quantitatively evaluate \name{} on several synthetic objects. \name{} achieves superior relighting accuracy compared to baseline inverse-rendering ~\cite{liang2023gs,saito2024relightable,jiang2024gaussianshader,li2025recap} and diffusion-based~\cite{jin2024neural_gaffer} methods, all while operating in under one second on a GPU. We further demonstrate its robustness through qualitative cross-scene object insertions on real-world captures under diverse lighting. Ultimately, despite relying on radially symmetric BRDF approximations and a low-pass SH basis, \name{} presents a highly practical trade-off, prioritizing instant, perceptually realistic relighting over the heavy computational costs of explicit modeling.

%In summary, our contributions are:
%\begin{itemize}
%\item \textbf{A fast (sub 10-second) relighting strategy for GS} that adapts each Gaussian's appearance to a target environment by analytically modulating SH coefficients,
%\item \textbf{A strategy to approximate environment maps} directly from GS models without additional supervision, and
%\item \textbf{State-of-the-art performance} on relighting benchmarks, demonstrating competitive but faster relighting compared to baseline optimization-based inverse rendering methods.
% \item \textbf{ Fast way of adding shadows
% An improved fine-grained object extraction pipeline for GS} that integrates object label weights into the scene fitting process to enable our object transfer application.
%\end{itemize}

%% file: sec/2_related.tex
\section{Related work}
\textbf{Neural-Based Inverse Rendering and Radiance Transfer.} Classical inverse rendering, the estimation of physical scene properties from images, relies on iterative optimizations using physics and geometry~\cite{boivin2001image,debevec2004estimating,durand2005frequency,marschner1998inverse}, often requiring controlled setups or known scene properties due to the ill-posed nature of the task. Modern inverse rendering approaches leverage deep learning and neural representations to decompose geometry, materials, and lighting from unstructured images. Early methods used feed-forward networks trained on large datasets to predict material attributes~\cite{sang2020single, li2020crowdsampling, yu2019inverserendernet}. Neural Radiance Fields (NeRFs)~\cite{barron2021mip,barron2022mipnerf360,mildenhall2021nerf} introduced per-scene optimization via MLP encodings, enabling methods like NeRF-in-the-Wild~\cite{martin2021nerf} to separate transient lighting from constant scene properties. Subsequent NeRF-based works integrated physically-based rendering to further disentangle object and lighting properties~\cite{srinivasan2021nerv, zhang2021nerfactor, yao2022neilf,zhang2023neilf++}. Recent advancements include TensoIR~\cite{Jin2023TensoIR}, which leverages tensor factorization to efficiently model illumination, and Neural-PBIR~\cite{sun2024neuralpbirreconstructionshapematerial}, a hybrid pipeline that pairs neural representations with explicit classical physics-based rendering.

In contrast, radiance transfer and ratio image methods~\cite{Mahajan2006, ramamoorthi2001signal} perform image-based relighting without explicit BRDF estimation by manipulating spherical harmonic (SH) identities. While these methods trade the ability to explicitly model complex, high-frequency material properties for computational efficiency, they provide a practical alternative to iterative optimization. \name{} brings these theoretical foundations into the Gaussian Splatting (GS) framework.

\textbf{Inverse Rendering and Relighting in GS.} Recent methods apply physically-based rendering to the 3D Gaussian Splatting (GS)~\cite{kerbl3Dgaussians} framework to estimate per-Gaussian surface properties~\cite{jiang2024gaussianshader,liang2023gs,R3DG2023,saito2024relightable}. Because single-environment captures inherently entangle lighting and materials, ReCap~\cite{li2025recap} requires cross-environment captures. Other works utilize precomputed transport: PRTGS~\cite{guo2024prtgsprecomputedradiancetransfer} adapts Precomputed Radiance Transfer but requires heavy precomputation for dense transport matrices, and LumiGauss~\cite{kaleta2024lumigaussrelightablegaussiansplatting} learns an SH-based illumination model for in-the-wild scenes but relies on explicit decomposition of material properties to relight. In contrast, \name{} leverages SH radiance transfer ratios~\cite{Mahajan2006, ramamoorthi2001signal,Troccoli2006RecoveringIA} to operate directly on existing, entangled SH coefficients. By formulating transfer as analytical irradiance and specularity ratios, \name{} bypasses explicit BRDF calculations and heavy precomputations entirely. While this trades explicit ultra-high-frequency material modeling for a radially symmetric approximation, it achieves near-instant updates over the costly and ill-posed optimization required by these baselines.

\textbf{Generative Relighting Priors for Neural Fields.} Another class of methods bypasses explicit inverse rendering using generative diffusion priors. IllumiNeRF~\cite{zhao2024illuminerf3drelightinginverse} and Neural Gaffer~\cite{jin2024neural_gaffer} employ a two-stage strategy: performing 2D image-space relighting conditioned on an environment map, then distilling the results into a 3D representation. While avoiding explicit material estimation, this paradigm has practical drawbacks. IllumiNeRF requires slow multi-view diffusion inference and full model retraining per scene, while Neural Gaffer is constrained to low-resolution ($256\times256$) outputs and can struggle with geometric consistency. Unlike these generative approaches, which rely on 2D priors prone to multi-view inconsistencies, \name{} uses a deterministic, physics-grounded SH radiance transfer. This mathematically principled approach ensures native 3D consistency and operates orders of magnitude faster.

% A distinct class of relighting methods similarly bypasses the 3DGS-IR problem by proposing alternative architectures or leveraging generative priors. For example, RNG~\cite{fan2025rng} replaces analytic shading models with a learned neural decoder (MLP) to handle non-surface-like geometries, such as fur or fabric, where explicit normals are ill-defined. GS³~\cite{bi2024gs3} introduces a novel 'triple splatting' pipeline, but is specialized for data captured from a light-stage. 
% \name{} achieves relighting analytically within the 3D Gaussian domain, producing instant, resolution-independent results without diffusion sampling or retraining. 
% This strategy offers a lightweight solution that shows visually appealing results in cross-scene transfer scenarios (see Fig.~\ref{fig:teaser}). 
% We propose a "reconstruct-once, relight-instantly" method that is fully analytical and operates as a lightweight post-processing step. Our method requires no new architectures, specialized data, generative priors, or costly retraining, as it directly modulates the SH coefficients of a standard pre-trained GS model.

%% file: sec/3_methods.tex
\section{Methods}
\begin{figure*}[t!]
\centering
\includegraphics[width=\linewidth]{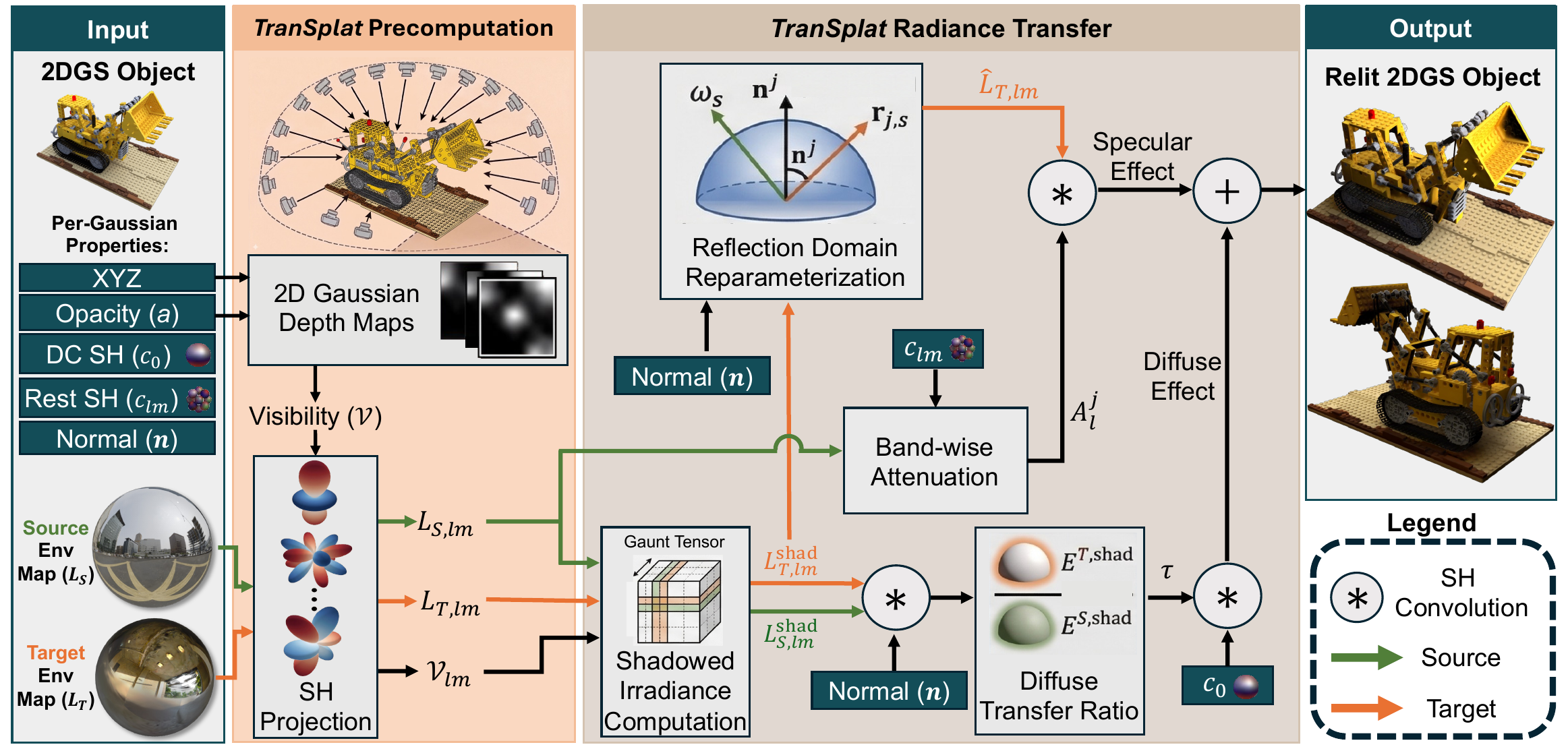}
\caption{\textbf{Overview of \name{}.} Given a pretrained 2DGS Object and source/target environment maps, \name{} performs four steps directly operating on Gaussian SH coefficients: \emph{(i)} \textbf{Precomputation} of a smooth visibility function from 2D Gaussian depth maps; \emph{(ii)} \textbf{SH-domain self-shadowing} via the Gaunt tensor to calculate shadowed incident radiance; \emph{(iii)} \textbf{Diffuse irradiance transfer}, computing a diffuse transfer ratio ($\tau$) applied to the DC SH coefficients ($c_0$); and \emph{(iv)} \textbf{Specularity-aware transfer}, utilizing reflection domain reparameterization and band-wise attenuation ($A_l^j$) applied to the higher-order SH coefficients ($c_{lm}$). No iterative optimization is required to output the relit 2DGS object.}
\label{fig:overview}
\end{figure*}

\name{} performs instant, object-centric relighting under arbitrary target illumination, bypassing explicit material decomposition. Given a pretrained 2D Gaussian Surfels (2DGS)~\cite{Huang2DGS2024} model of an object and source \& target environment maps $L_S$ \& $L_T$, we first precompute per-surfel visibility maps as a function of viewpoint. We then directly transfer lighting to the object by modulating the spherical harmonics (SH) coefficients of its constituent surfels to match the target environment $L_T$. \name{} consists of three core components: diffuse transfer via irradiance ratios, specular transfer via reflection-domain band-wise attenuation, and self-shadowing via the Gaunt tensor. We illustrate the complete framework in Fig.~\ref{fig:overview}.

\subsection{Radiance Transfer Background}\label{sec:background}
We build on the theoretical framework for radiance transfer with spherical harmonics from~\cite{basri2003lambertian, ramamoorthi2001signal}, drawing on ratio and quotient image methods~\cite{Troccoli2006RecoveringIA}. Consider a point on a 3D surface with outward normal $\mathbf{n}$. The reflection equation~\cite{kajiya1986rendering} relates outgoing radiance in direction $\boldsymbol{\omega}_o$ to the incident illumination:
\begin{equation}
B(\mathbf{n}, \boldsymbol{\omega}_o) = \int_{S^2} L(\boldsymbol{\omega}_i)\, \rho(\boldsymbol{\omega}_i, \boldsymbol{\omega}_o)\, (\boldsymbol{\omega}_i \cdot \mathbf{n})_+ \, d\boldsymbol{\omega}_i,
\label{eq:rendering}
\end{equation}
where $L(\boldsymbol{\omega}_i)$ is the incoming radiance, $\rho(\boldsymbol{\omega}_i, \boldsymbol{\omega}_o)$ is the BRDF, and $(\boldsymbol{\omega}_i \cdot \mathbf{n})_+$ is the clamped cosine, with $(\cdot)_+ = \max(\cdot,0)$.
\noindent\textbf{Diffuse (Lambertian) case.}
For a Lambertian surface with constant albedo $\rho_d$, the BRDF reduces to $\rho(\boldsymbol{\omega}_i, \boldsymbol{\omega}_o) = \rho_d/\pi$, which can be factored from the integral:
\begin{equation}
B_d(\mathbf{n}) = \frac{\rho_d}{\pi} \int_{S^2} L(\boldsymbol{\omega}_i)\, (\boldsymbol{\omega}_i \cdot \mathbf{n})_+\, d\boldsymbol{\omega}_i = \frac{\rho_d}{\pi}\, E(\mathbf{n}),
\label{eq:lambertian}
\end{equation}
where $E(\mathbf{n})$ is the irradiance at normal $\mathbf{n}$. The irradiance integral is a spherical convolution of $L$ with the clamped cosine, which in the SH basis yields:
\begin{equation}
E(\mathbf{n}) = \sum_{l=0}^{l_{\max}} \sum_{m=-l}^{l} A^{\text{Lamb}}_l\, L_{lm}\, Y_{lm}(\mathbf{n}),
\label{eq:irradiance_sh}
\end{equation}
where $A^{\text{Lamb}}_l$ are the analytic Zonal Harmonic (ZH) coefficients of the clamped cosine ($A^{\text{Lamb}}_0{=}\pi,\; A^{\text{Lamb}}_1{=}2\pi/3,\; A^{\text{Lamb}}_2{=}\pi/4$, and $A^{\text{Lamb}}_l{=}0$ for odd $l \geq 3$). Because $\rho_d$ is an intrinsic material constant invariant across environments, it cancels in irradiance ratios -- a crucial fact underpinning our radiance transfer strategy.

\textbf{Specular and view-dependent case.}
For non-Lambertian appearance, the outgoing radiance depends on both $\mathbf{n}$ and $\boldsymbol{\omega}_o$. Unlike the Lambertian case, lighting changes cannot be expressed as a single normal-dependent ratio. According to the SH Convolution Theorem~\cite{ramamoorthi2001signal}, the reflected radiance of a glossy surface is equivalent to filtering the incident illumination with the material's BRDF kernel. We leverage this property in Sec.~\ref{sec:specular_transfer} to extract an approximate BRDF attenuation profile from the pretrained Gaussians and modulate the higher-order SH bands directly without explicit modeling of microfacet parameters. Because the GS coefficients entangle illumination and material, this extraction is a heuristic rather than a theoretically exact inversion.

\subsection{Object Representation and Precomputation}\label{sec:precomputation}
We represent a 3D object using the 2D Gaussian Surfels (2DGS)~\cite{Huang2DGS2024} framework. Surfels provide a flat structure that yields better high-quality surface normals essential for accurate irradiance computation compared to standard 3D Gaussians~\cite{kerbl3Dgaussians}. We further assume that incident source radiance $L_S(\theta, \phi)$ and target radiance $L_T(\theta, \phi)$ are given as spatial HDRI environment maps. %We can also estimate $L_S$ and $L_T$ directly from the trained source and target GS scene representations by rendering incoming radiance from the object's center into an equirectangular map and utilizing an off-the-shelf LDR-to-HDR model~\cite{liu2020single} to recover dynamic range (see Supplementary).

GS models commonly capture view-dependent appearances using SH coefficients up to degree $l_{\max}=3$. Hence, we also project both $L_S$ and $L_T$ into the SH domain using these same basis functions, yielding the unshadowed incident radiance SH coefficients $L_{S,lm}$ and $L_{T,lm}$. Our experiments suggest that including SH coefficients beyond $l{=}3$ yields minimal relighting performance gains (Table.~\ref{tab:sh_order_ablation}). Thus, restricting our projection to $l{=}3$ maintains both real-time performance and plug-and-play usage of standard 2DGS-trained assets.

Next, we precompute the object's self-occlusion profile under the source illumination $L_S$. For each surfel at $\mathbf{x}$, we sample uniform directions to define a visibility function $\mathcal{V}(\mathbf{x}, \boldsymbol{\omega})$ based on the depth maps already computed by the 2DGS rasterizer. We observed that projecting a discontinuous binary visibility function onto the SH basis causes Gibbs ringing artifacts. To mitigate this, we replace the binary threshold with Exponential Shadow Maps (ESM) coupled with an 8-tap Poisson disk filter to yield a smooth function. Projecting these smooth visibility samples onto the SH basis yields coefficients $\mathcal{V}_{lm}(\mathbf{x})$. Additional implementation details for visibility and shadow-map precomputation are provided in Supplementary Sec. 2.

\subsection{Self-Shadowing via Gaunt Tensor}\label{sec:shadows}
\begin{figure}[t!]
\centering
\includegraphics[width=\linewidth]{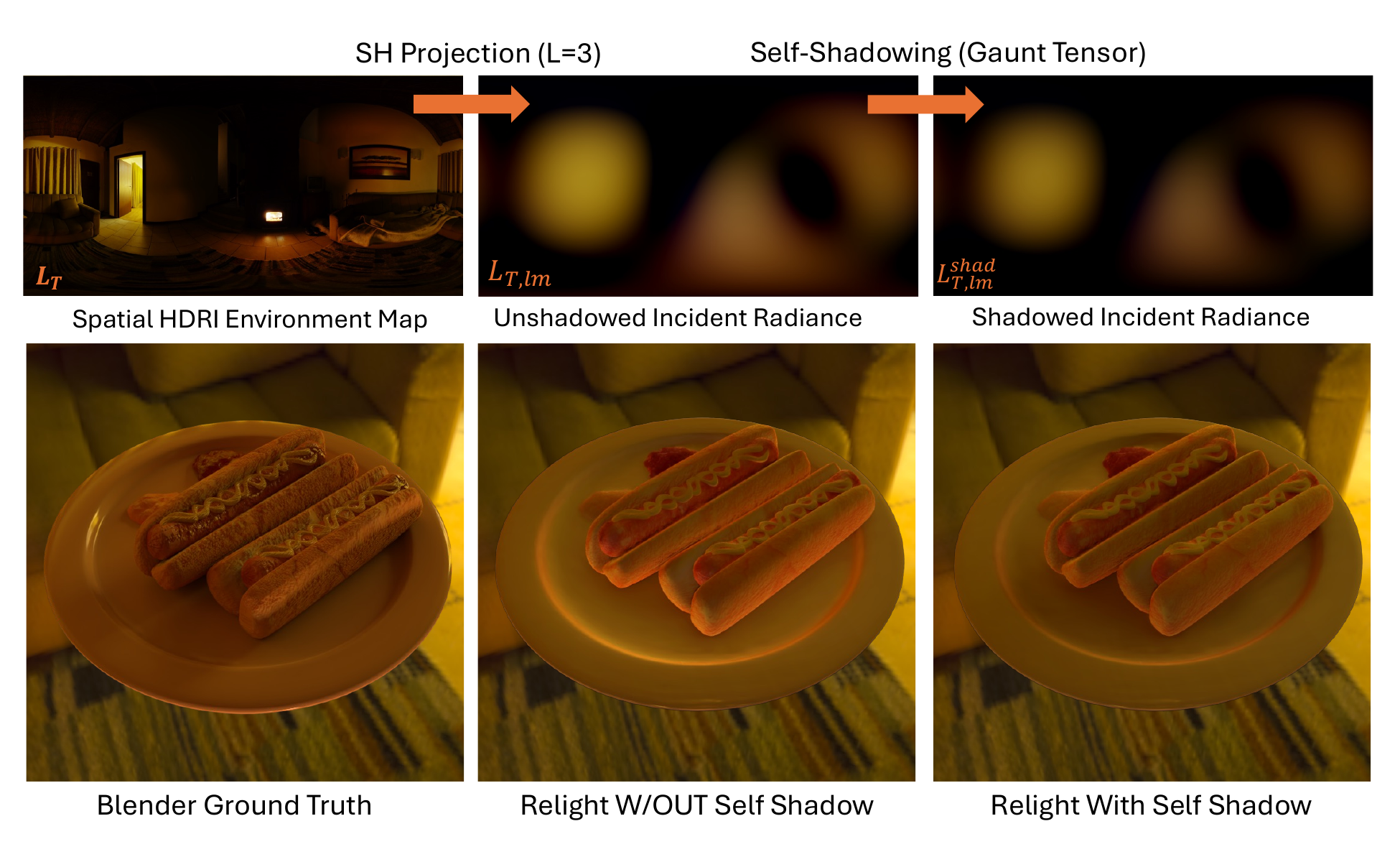}
\caption{\textbf{Environment Map SH projection and self-shadow visualization.}
\emph{Top row:} Original HDR environment map (left), its band-limited $l{\le}3$ SH projection representing the unshadowed incident radiance $L_{lm}$ (center), and the per-surfel shadowed incident radiance $L_{lm}^{\text{shad}}$ produced by the Gaunt tensor (Eq.~\ref{eq:gaunt_tensor}) (right). The SH projection smooths out high-frequency content, while the Gaunt tensor further attenuates incoming radiance from self-occluded directions.
\emph{Bottom row:} Impact on relighting the \emph{Hotdog} object. Using unshadowed $L_{lm}$ yields a flat, shadow-free appearance; applying $L_{lm}^{\text{shad}}$ produces physically consistent self-cast shadows that closely match the Blender ground truth.}
\label{fig:envmap_visualization}
\end{figure}

To ensure physically realistic self-occlusion under the new target illumination $L_T$, we compute shadowing within the SH frequency domain, akin to Precomputed Radiance Transfer (PRT~\cite{Sloannetal2002}). With the precomputed local occlusion coefficients $\mathcal{V}_{lm}(\mathbf{x})$ and the unshadowed incident radiance SH coefficients ($L_{S,lm}$ and $L_{T,lm}$), we compute the shadowed incident radiance for both the source and target via the Gaunt tensor:
\begin{equation}
L^{\text{shad}}_{X,k}(\mathbf{x}) = \sum_{p,q} \mathcal{G}_{kpq}\, L_{X,p}\, \mathcal{V}_q(\mathbf{x}), \quad \text{for } X \in \{S, T\},
\label{eq:gaunt_tensor}
\end{equation}
where $p, q, k$ denote flattened SH indices representing $(l,m)$ pairs (distinct from the Gaussian index $j$ used elsewhere), and $\mathcal{G}_{kpq}$ encodes the integral of three SH basis functions. This formulation yields our final shadowed incident radiance SH coefficients, $L_{S,lm}^{\text{shad}}$ and $L_{T,lm}^{\text{shad}}$, visually demonstrated in Fig.~\ref{fig:envmap_visualization}.

\subsection{Diffuse Radiance Transfer}\label{sec:diffuse_transfer}
\name{} applies a radiance transfer function to the base color of the object. Let $c_{lm}^j$ denote the spherical harmonic coefficients for the $j$-th Gaussian surfel. We make the practical assumption that the DC ($l{=}0$) component, $c_0^j \equiv c_{00}^j$ (the unique $l{=}m{=}0$ coefficient), primarily captures the diffuse, view-independent color. Strictly, $c_0^j$ encodes the spherical mean of the full radiance field and can carry a DC contribution from any specular lobe; this approximation is valid when the diffuse term dominates the spherical mean, which holds for most non-mirror surfaces. Standard GS optimizes these coefficients to mimic photometric appearance, deeply entangling illumination and material.

Standard irradiance evaluation relies on the unshadowed SH coefficients. However, to incorporate view-independent self-occlusion, we compute the per-surfel diffuse transfer ratio utilizing the shadowed incident radiance. By substituting our shadowed incident radiance coefficients $L_{S,lm}^{\text{shad}}$ and $L_{T,lm}^{\text{shad}}$ into the standard irradiance convolution (Eq.~\ref{eq:irradiance_sh}), we obtain the shadowed irradiance $E^{S,\text{shad}}(\mathbf{n}^j)$ and $E^{T,\text{shad}}(\mathbf{n}^j)$. The per-surfel diffuse transfer ratio is then:
\begin{equation}
\tau^j = \frac{E^{T,\text{shad}}(\mathbf{n}^j)}{E^{S,\text{shad}}(\mathbf{n}^j) + \varepsilon_d},
\label{eq:transfer_ratio}
\end{equation}
where $\varepsilon_d$ provides numerical stability. We update the view-independent DC coefficient $c_0^j$ via a scalar multiplication:
\begin{equation}
c_{0}^{j,T} = \tau^j \cdot c_{0}^{j}.
\label{eq:dc_update}
\end{equation}

\subsection{Specularity-Aware Radiance Transfer via SH Deconvolution}\label{sec:specular_transfer}
Unlike the Lambertian case, the higher-order SH bands ($l \geq 1$) encode view-dependent, specular appearance. The key distinction is that a specular BRDF $\rho(\theta_r)$ is radially symmetric around the \emph{reflection direction} $\mathbf{r}(\boldsymbol{\omega}_o)$, not the surface normals. Thus, the SH Convolution Theorem must be applied in the reflection-parameterized domain, not the normal-aligned frame used for diffuse transfer.

\begin{figure*}[t!]
\centering
\includegraphics[width=1\linewidth]{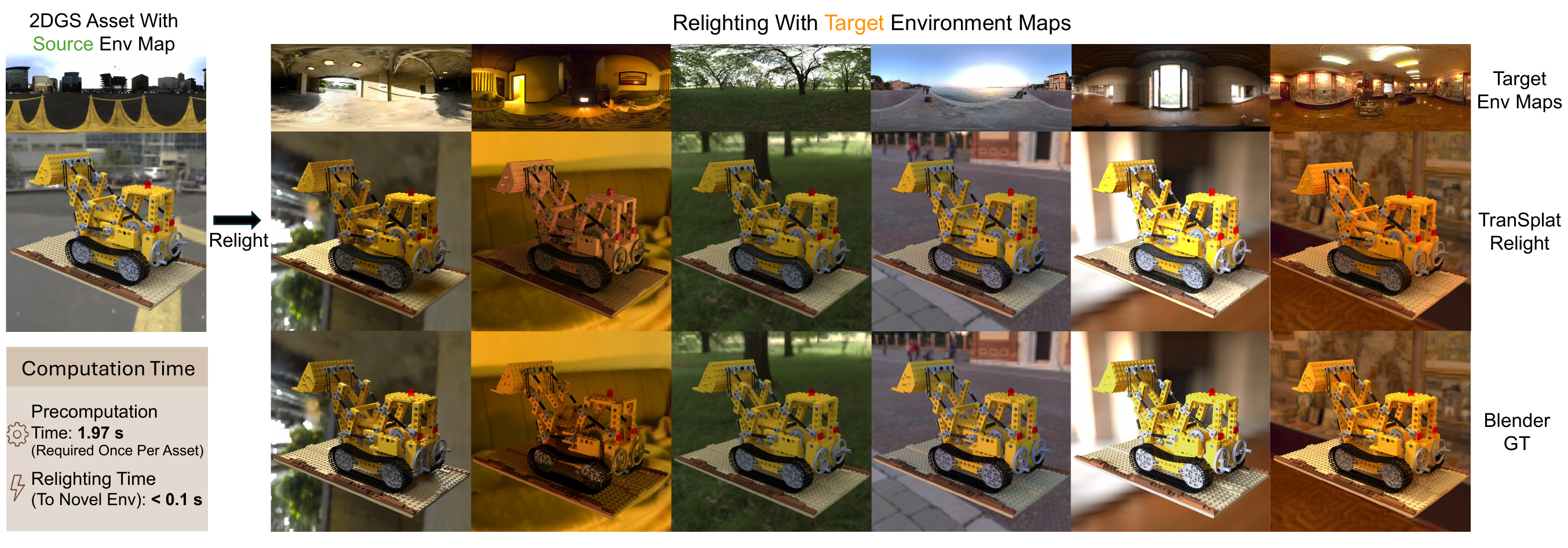}
\caption{\textbf{Sample \name{} results in relighting a synthetic object from a source environment (\emph{City}, left) to novel target environments.} \name{} accurately reproduces complex shading effects and shadows, closely matching the physically-based Blender ground truth (bottom row) across a wide variety of target illuminations.}
\label{fig:relight_visual}
\end{figure*}

\noindent\textbf{Reflection-Domain Reparameterization.}
For a given view direction $\boldsymbol{\omega}_o$, the pure mirror-reflection direction about normal $\mathbf{n}^j$ is:
\begin{equation}
\mathbf{r}^j(\boldsymbol{\omega}_o) = \boldsymbol{\omega}_o - 2(\boldsymbol{\omega}_o \cdot \mathbf{n}^j)\,\mathbf{n}^j.
\label{eq:reflect_dir}
\end{equation}
In the local frame with $\mathbf{r}^j$ as its z-axis, the specular BRDF becomes a zonal harmonic. The SH Convolution Theorem~\cite{ramamoorthi2001signal} then gives the reflected radiance at Gaussian $j$, expressed in this reflection frame, as a per-band product:
\begin{equation}
c_{lm}^j \approx \hat{c}_l \cdot \rho_l \cdot L_{S,lm}^{\text{refl}},
\label{eq:sh_convolution}
\end{equation}
where $\hat{c}_l = \sqrt{4\pi/(2l+1)}$, $\rho_l$ are the ZH coefficients of the specular BRDF $\rho(\theta_r)$, and $L_{S,lm}^{\text{refl}}$ are the unshadowed source incident radiance SH coefficients ($L_{S,lm}$) evaluated in the reflection frame.

\textbf{Analytic Bandwise Attenuation.}
Eq.~\ref{eq:sh_convolution} shows that the band-$l$ energy of the stored Gaussian coefficients encodes the product of the BRDF's frequency response and the source environment's energy at that band. To isolate the BRDF profile, we leverage the fact that SH band energy $\|\mathbf{c}_l^j\|^2 = \sum_m |c_{lm}^j|^2$ is \emph{rotationally invariant}—it is the same regardless of which coordinate frame the coefficients are expressed in. This allows us to form the ratio directly from the globally-stored $c_{lm}^j$ and the unshadowed source incident radiance SH coefficients $L_{S,lm}$ without explicitly rotating into the per-surfel reflection frame:
\begin{equation}
A_l^j = \frac{\|\mathbf{c}_{l}^j\|}{\|\mathbf{L}_{S,l}\| + \varepsilon_s} \approx \hat{c}_l \cdot \rho_l, \qquad 1 \leq l \leq l_{\max}.
\label{eq:al_extraction}
\end{equation}
The constant $\hat{c}_l$ is absorbed into $A_l^j$, yielding an effective, per-band attenuation profile. Three approximations are implicit in this step: \emph{(i)} the GS optimization entangles illumination and material, so $c_{lm}^j$ does not purely encode the BRDF response; \emph{(ii)} the band-norm ratio in Eq.~\ref{eq:al_extraction} recovers only the magnitude of the per-band response, discarding within-band phase information; and \emph{(iii)} the low SH order ($l_{\max}{=}3$) acts as a low-pass filter, limiting the accuracy with which sharp specular lobes can be represented. Under these approximations, $\rho_l$ cannot be recovered exactly; rather, $A_l^j$ serves as a data-driven proxy that absorbs the net effect of the material's frequency response as seen through the pretrained model. Since this proxy is learned under the source illumination and is independent of the environment, it transfers directly to the target illumination as an effective BRDF attenuation profile.

\textbf{Target Transfer.} To construct the new specular coefficients directly in the global viewing frame without requiring complex inverse SH rotations, we perform the transfer via a sampling-and-projection scheme. We uniformly sample a set of global viewing directions $\{\boldsymbol{\omega}_s\}$, compute their corresponding reflection directions $\mathbf{r}^j(\boldsymbol{\omega}_s)$ about the surfel normal $\mathbf{n}^j$ via Eq.~\ref{eq:reflect_dir}, and evaluate the shadowed target SH representation $L_{T,lm}^{\text{shad}}$ at these pure reflection directions. Because these evaluated radiance samples are intrinsically mapped to the global viewing directions $\{\boldsymbol{\omega}_s\}$, projecting them back into the SH basis yields the unattenuated target specular coefficients $\hat{L}_{lm}^{j,T}$ natively in the global view frame. Finally, applying the extracted, rotationally invariant attenuation profile approximates the band-wise BRDF filtering under the new environment:
\begin{equation}
c_{lm}^{j,T} = A_l^j \cdot \hat{L}_{lm}^{j, T}, \qquad l \geq 1.
\label{eq:specular_transfer}
\end{equation}
Because this projection naturally handles coordinate frame alignment, no further local-to-global transformations are required, allowing for accurate reproduction of complex shading effects across novel environments (see Fig.~\ref{fig:relight_visual} for qualitative results).

%% file: sec/4_experiments.tex
\section{Experiments}
We qualitatively and quantitatively evaluated \name{} on six synthetic objects from the TensoIR and Blender datasets~\cite{Jin2023TensoIR,stojanov2021using_toys4k}, using relighting accuracy (PSNR, SSIM~\cite{wang2004image}, LPIPS~\cite{zhang2018unreasonable}), runtime, and peak VRAM usage (GB) as metrics. We compared \name{} against five recent baselines, specifically prioritizing methods capable of real-time rendering once the object is relit. To this end, we included several 3DGS-based approaches utilizing explicit physically-based rendering (PBR) decomposition (GS-IR~\cite{liang2023gs}, R3DGS~\cite{R3DG2023}, GaussianShader~\cite{jiang2024gaussianshader}, and Recap~\cite{li2025recap}). We excluded traditional NeRF-based inverse rendering methods (e.g., TensoIR~\cite{Jin2023TensoIR}), as their slow volumetric rendering precludes real-time application. Finally, we included Neural Gaffer~\cite{jin2024neural_gaffer}, a representative of the emerging diffusion-guided 3D distillation approaches, which leverages a pretrained 2D diffusion model for relighting inference and then distills the results into a TensoRF \cite{chen2022tensorftensorialradiancefields} representation. We conducted all experiments on a single NVIDIA A6000 GPU.

\begin{figure*}[t!]
  \centering
  \includegraphics[width=1\linewidth]{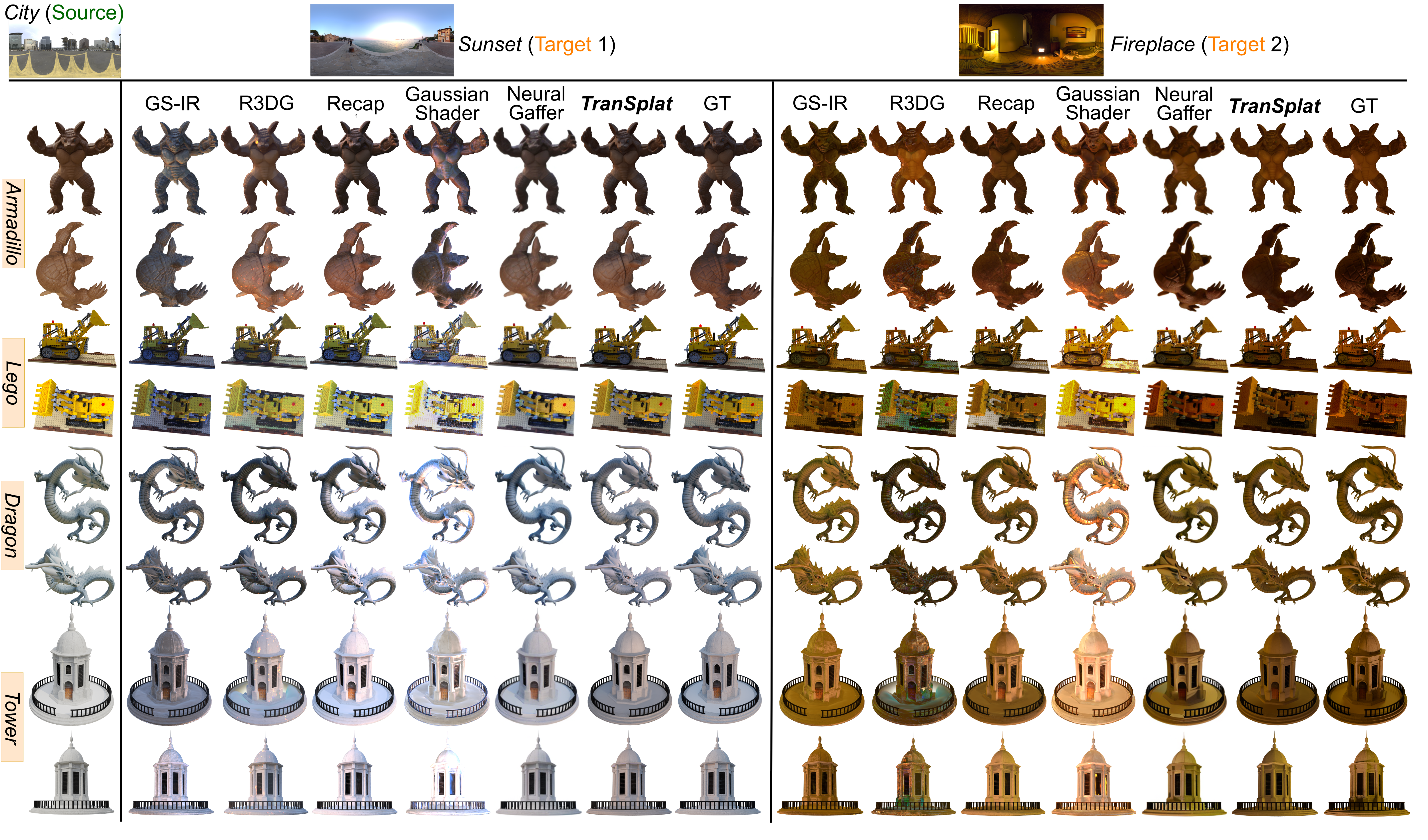}
  \caption{\textbf{Relighting results using two Blender-generated (\emph{Dragon}~\cite{stojanov2021using_toys4k} and \emph{Tower}) and two TensoIR~\cite{Jin2023TensoIR} objects.} We apply the same HDR source environment map (\emph{City}) to all objects, and then relight them to three other environments: \emph{Sunset}, \emph{Fireplace}, and \emph{Forest}. \name{} achieves the best visual quality in terms of color and shading effects compared to ground truth (GT). See Table~\ref{tab:comparison_extended} for corresponding metrics.}
  \label{fig:relighting_results}
\end{figure*}

\begin{table*}[t!]
  \centering
  \renewcommand{\arraystretch}{1.15}
  \setlength{\tabcolsep}{4pt}
  \caption{\textbf{Quantitative evaluation of relighting methods.} We compare \name{} against GS-IR~\cite{liang2023gs}, GaussianShader~\cite{jiang2024gaussianshader}, R3DGS~\cite{R3DG2023}, Recap~\cite{li2025recap}, and Neural~Gaffer~\cite{jin2024neural_gaffer} on three environment maps (\emph{Fireplace}, \emph{Forest}, \emph{Sunset}) using standard image reconstruction metrics (PSNR, SSIM, LPIPS). We highlight the best results in \best{green}, and second-best in \second{yellow}. \name{} performs the best of all methods in almost every scenario.}
  \label{tab:comparison_extended}
  \resizebox{\textwidth}{!}{
  \begin{tabular}{l c c c c c c | c c c c c c | c c c c c c}
  & \multicolumn{6}{c}{\textbf{\emph{Fireplace}}}
  & \multicolumn{6}{c}{\textbf{\emph{Forest}}}
  & \multicolumn{6}{c}{\textbf{\emph{Sunset}}} \\
  \cmidrule(lr){2-7}\cmidrule(lr){8-13}\cmidrule(lr){14-19}
  
  & GS-IR & GShader & R3DGS & Recap & Neural Gaffer & \textbf{\name{}}
  & GS-IR & GShader & R3DGS & Recap & Neural Gaffer & \textbf{\name{}}
  & GS-IR & GShader & R3DGS & Recap & Neural Gaffer & \textbf{\name{}} \\
  
  \midrule
  \multicolumn{19}{c}{\textbf{PSNR (dB) $\uparrow$}} \\
  \midrule
  \emph{Lego} & \second{23.99} & 12.74 & 22.83 & 19.39 & 23.79 & \best{30.09}
       & 23.04 & 13.77 & \second{26.51} & 23.43 & 23.86 & \best{32.00}
       & 24.09 & 12.73 & \second{26.00} & 22.40 & 23.73 & \best{31.04} \\
  \emph{Armadillo} & 27.81 & 22.66 & 27.07 & \second{30.83} & 30.16 & \best{33.57}
            & 26.99 & 21.97 & 26.02 & \second{30.71} & 29.14 & \best{34.55}
            & 27.49 & 20.29 & 25.72 & \second{29.99} & 28.90 & \best{33.84} \\
  \emph{Dragon} & 23.89 & 15.96 & 23.59 & \second{24.49} & 23.28 & \best{26.94}
         & 21.04 & 16.08 & 24.44 & 25.24 & \second{26.72} & \best{29.77}
         & 22.99 & 15.55 & 23.10 & 23.74 & \second{27.10} & \best{29.46} \\
  \emph{Ficus} & 26.96 & 25.36 & 24.87 & \second{28.10} & 24.40 & \best{28.92}
        & 26.23 & 22.74 & 27.22 & \second{31.40} & 25.44 & \best{33.66}
        & 27.44 & 22.12 & 28.50 & \second{31.24} & 26.09 & \best{32.23} \\
  \emph{Tower} & 22.60 & 13.92 & 21.98 & 24.32 & \second{24.77} & \best{27.12}
        & 18.66 & 13.77 & 24.09 & 23.66 & \second{25.60} & \best{29.69}
        & 20.32 & 13.78 & 25.05 & 23.64 & \second{26.04} & \best{29.10} \\
  \emph{Hotdog} & \second{24.06} & 10.87 & 22.51 & 13.19 & 23.58 & \best{28.15}
         & 19.03 & 15.92 & \second{25.71} & 19.26 & 21.25 & \best{27.97}
         & 20.31 & 14.84 & \second{25.87} & 18.76 & 21.45 & \best{27.51} \\
  \midrule
  \multicolumn{19}{c}{\textbf{SSIM} $\uparrow$} \\
  \midrule
  \emph{Lego} & \second{0.86} & 0.75 & 0.82 & 0.83 & 0.84 & \best{0.93}
       & 0.87 & 0.82 & \second{0.93} & 0.92 & 0.85 & \best{0.96}
       & 0.89 & 0.80 & \second{0.92} & \second{0.92} & 0.85 & \best{0.95} \\
  \emph{Armadillo} & 0.92 & 0.89 & 0.92 & 0.93 & \second{0.95} & \best{0.96}
            & 0.93 & 0.92 & 0.95 & \second{0.97} & 0.96 & \best{0.98}
            & 0.94 & 0.91 & 0.96 & \second{0.97} & 0.96 & \best{0.98} \\
  \emph{Dragon} & 0.88 & 0.83 & 0.87 & \second{0.89} & 0.88 & \best{0.91}
         & 0.89 & 0.87 & 0.94 & \second{0.95} & 0.92 & \best{0.97}
         & 0.90 & 0.87 & 0.94 & \second{0.95} & 0.92 & \best{0.97} \\
  \emph{Ficus} & \second{0.96} & 0.95 & 0.95 & \best{0.97} & 0.95 & \best{0.97}
        & 0.96 & 0.95 & 0.97 & \second{0.98} & 0.95 & \best{0.99}
        & 0.96 & 0.94 & \second{0.97} & \best{0.98} & 0.95 & \best{0.98} \\
  \emph{Tower} & 0.87 & 0.82 & 0.85 & \second{0.89} & 0.88 & \best{0.91}
        & 0.85 & 0.82 & 0.90 & \best{0.95} & \second{0.92} & \best{0.95}
        & 0.90 & 0.85 & 0.93 & \second{0.96} & 0.93 & \best{0.97} \\
  \emph{Hotdog} & \second{0.88} & 0.76 & 0.85 & 0.79 & \second{0.88} & \best{0.93}
         & 0.89 & 0.88 & \second{0.93} & 0.90 & 0.88 & \best{0.95}
         & 0.91 & 0.88 & \second{0.94} & 0.90 & 0.90 & \best{0.95} \\
  \midrule
  \multicolumn{19}{c}{\textbf{LPIPS} $\downarrow$} \\
  \midrule
  \emph{Lego} & \second{0.10} & 0.15 & 0.12 & \second{0.10} & 0.15 & \best{0.07}
       & 0.09 & 0.12 & \second{0.05} & 0.06 & 0.15 & \best{0.04}
       & 0.08 & 0.13 & \second{0.06} & \second{0.06} & 0.16 & \best{0.04} \\
  \emph{Armadillo} & 0.07 & 0.07 & \second{0.06} & \best{0.05} & 0.07 & \best{0.05}
            & \second{0.06} & 0.07 & \best{0.04} & \best{0.04} & 0.07 & \best{0.04}
            & 0.06 & 0.07 & 0.05 & \best{0.03} & 0.06 & \second{0.04} \\
  \emph{Dragon} & 0.08 & 0.13 & 0.10 & \second{0.07} & 0.11 & \best{0.06}
         & 0.08 & 0.10 & 0.06 & \second{0.05} & 0.09 & \best{0.03}
         & 0.08 & 0.11 & 0.06 & \second{0.05} & 0.10 & \best{0.04} \\
  \emph{Ficus} & \second{0.03} & 0.04 & 0.04 & \second{0.03} & 0.05 & \best{0.02}
        & 0.03 & 0.04 & \second{0.02} & \best{0.01} & 0.05 & \best{0.01}
        & 0.03 & 0.05 & 0.03 & \best{0.01} & 0.05 & \second{0.02} \\
  \emph{Tower} & \second{0.08} & 0.11 & 0.12 & \best{0.06} & 0.10 & \best{0.06}
        & 0.09 & 0.13 & 0.08 & \best{0.04} & 0.09 & \second{0.05}
        & \second{0.09} & 0.14 & \second{0.09} & \best{0.04} & \second{0.09} & \best{0.04} \\
  \emph{Hotdog} & \second{0.13} & 0.15 & 0.15 & 0.14 & 0.15 & \best{0.09}
         & 0.14 & 0.13 & \second{0.11} & \second{0.11} & 0.15 & \best{0.09}
         & 0.12 & 0.15 & \second{0.11} & 0.12 & 0.15 & \best{0.09} \\
  \bottomrule
  \end{tabular}
  }
  \end{table*}

\subsection{Relighting Accuracy Evaluation}
\label{sec:relighting_results}
We first compare \name{} against all baseline relighting algorithms. We used six objects (\emph{Armadillo}, \emph{Lego}, \emph{Ficus}, \emph{Dragon}, \emph{Tower}, \emph{Hotdog}) and HDR maps commonly used by prior works. We obtained ground truth relighting results for \emph{Dragon} and \emph{Tower} in Blender. For a consistent comparison, we used the same source environment (\emph{City}) for all cases. Table~\ref{tab:comparison_extended} presents quantitative results, and  Fig.~\ref{fig:relighting_results} presents qualitative rendering outputs. \name{} achieves the best overall accuracy, ranking first in PSNR, SSIM, and LPIPS on the majority of combinations of all six objects and three target environments. All methods relight objects to the correct color tone, but the results from GaussianShader~\cite{jiang2024gaussianshader} appear noticeably brighter. Neural Gaffer~\cite{jin2024neural_gaffer} produces accurate shading and shadows, but its output resolution is constrained to 256{\(\times\)}256 pixels.

\subsection{Object Insertion And Relighting into Scenes}
\begin{figure*}[t!]
  \centering
  \includegraphics[width=0.95\linewidth]{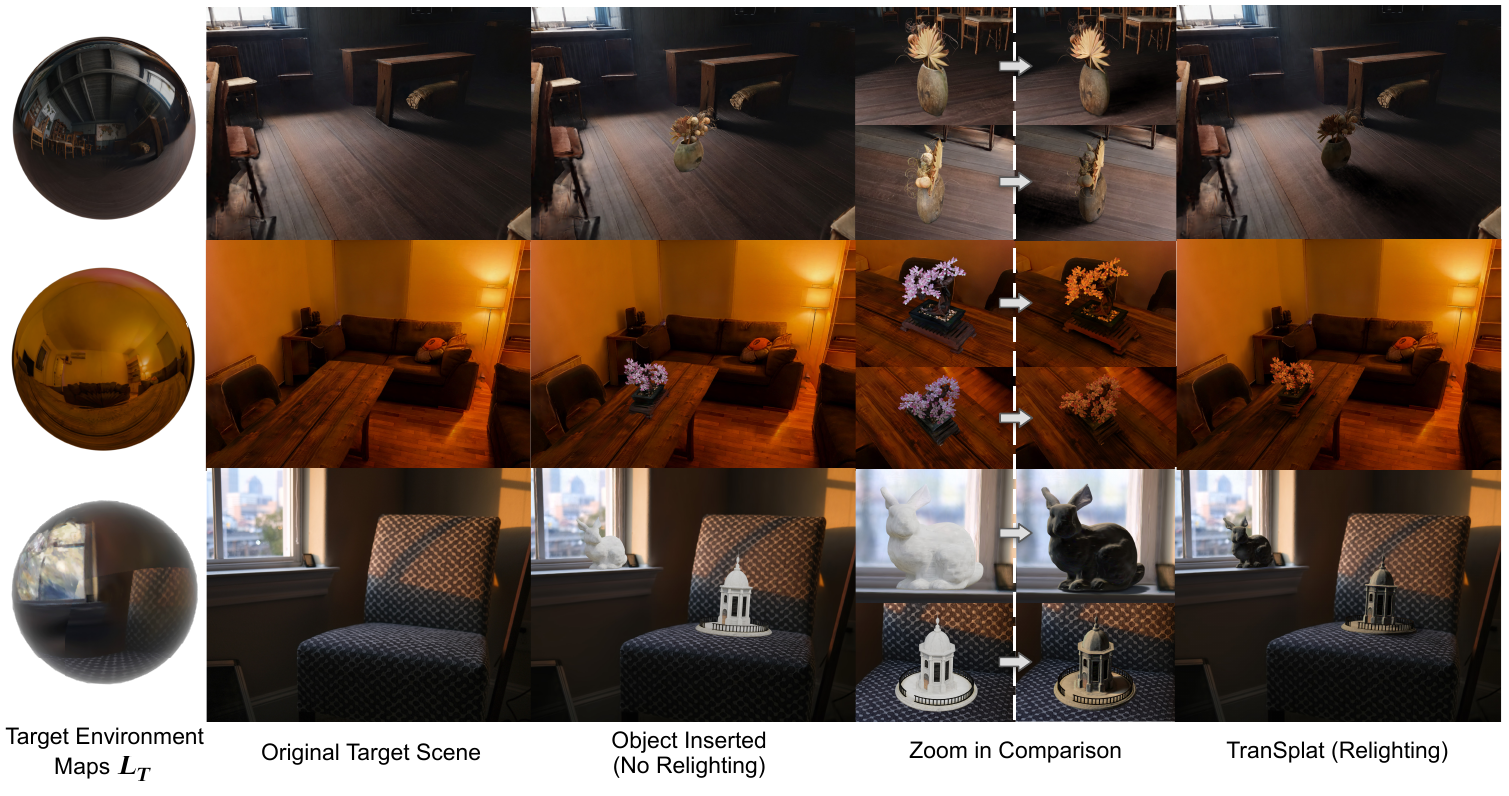} 
  \vspace{-0.15in}
  \caption{\textbf{Sample results demonstrating TranSplat’s object insertion and relighting in real scenes.} Inserting objects into another scene without relighting yields inconsistent appearances. However, using \name{}, renderings appear more natural with accurate shading and shadows. Details are best observed in videos on the project website.
 }
  \label{fig:main_results}
\end{figure*}

Next, we demonstrate using \name{} to relight pretrained 2DGS assets for a full cross-scene insertion task. We prepared three target environments with distinct lighting (see Fig.~\ref{fig:main_results}). We generated the first two scenes using Marble~\cite{worldlabs_marble_software}, and captured the last scene ourselves with a camera. We inserted two synthetic Blender objects (\emph{Bunny} and \emph{Tower}), and two real objects (\emph{Vase} and \emph{Bonsai}) from MipNerf360 (\emph{Kitchen} and \emph{Bonsai} scenes)~\cite{barron2022mipnerf360}. \emph{Bunny} and \emph{Tower} use a \emph{City} environment source map, and we estimated all other source and target lighting via cube map sampling from the trained representation (see Supplementary, Sec. 6 for details). We added ground shadows to enhance realism by treating all Gaussians near the object as also belonging to the object during the self-shadowing computation (Sec.~\ref{sec:shadows}). We compare objects inserted using \name{} before and after relighting in Fig.~\ref{fig:main_results}. Directly inserting objects into the target scene without relighting results in unrealistic appearances. \name{} adjusts both general brightness along with realistic shading effects, such as the objects appearing brighter on the side facing the window in the first and second rows. 

%Because \name{} relights any pre-trained 2DGS asset given only source and target environment maps, it integrates naturally as a post-processing step in a full cross-scene insertion pipeline. To demonstrate this, 

\begin{table*}[t!]
\centering
\small
\renewcommand{\arraystretch}{1.15}
\setlength{\tabcolsep}{5pt}
\caption{\textbf{Comparison of efficiency and requirements of all methods.} For each algorithm, we present its strategy, support for pretrained GS assets, scene modeling precomputation time, relighting time, and peak VRAM usage. Recap and GaussianShader do not have separate precomputation and relighting phases, so we report single values for them.
% \guha{This caption doesn't really say what the table is presenting. Just say something like: For each algorithm, we present its strategy, ability to use pre-trained GS assets, scene modeling precomputation time, relighitng time, and peak VRAM usage.} \guha{The ``Scene Modeling Time + Precomputation'' and ``Time to Relight (To Novel Env.)'' are confusing. Just say ``Scene Modeling Precomputation Time'' and ``Relighting Time''.} 
Unlike baselines that require costly explicit material decomposition or hours of iterative inference per scene, \name{} operates directly on pre-trained vanilla 2D Gaussian assets. By fully decoupling the base geometry optimization from the illumination transfer, our method reduces novel-environment relighting to a sub-second analytical pass.}
\label{tab:timing_transplat_compact}
\begin{tabular}{@{}llcccc@{}}
  \toprule
  \textbf{Method} & \textbf{Paradigm} & \textbf{\makecell[c]{Relights Pre-Trained\\ GS Assets?}} & \textbf{\makecell[c]{Scene Modeling\\Precomputation Time}} & \textbf{\makecell[c]{Relighting Time}} & \textbf{\makecell[c]{Peak GPU\\VRAM}} \\
  \midrule

  \multicolumn{6}{@{}l}{\textit{3DGS Inverse Rendering}} \\
  GS-IR~\cite{liang2023gs} & Explicit Decomp. & No & 16 min & 47 min & 2.59 GB \\
  Relightable 3DGS~\cite{R3DG2023} & Explicit Decomp. & No & 16 min & 84 min  & 11.26 GB \\
  GaussianShader~\cite{jiang2024gaussianshader} & Joint Optimization & No & \multicolumn{2}{c}{97 min} & 8.58 GB \\
  Recap~\cite{li2025recap} & Joint Optimization & No & \multicolumn{2}{c}{60 min} & 1.81 GB \\
  \addlinespace

  \multicolumn{6}{@{}l}{\textit{Generative / Diffusion}} \\
  Neural Gaffer~\cite{jin2024neural_gaffer} & \makecell[l]{Diffusion-Guided\\3D Distillation} & No & 23 min  & 15 min  & 4.95 GB \\
  \midrule

  \rowcolor{blue!5}
  \textbf{\name{} (Ours)} & \textbf{Radiance Transfer} & \textbf{Yes} & \textbf{$\sim$7 min}\textsuperscript{\textdagger} & \textbf{$\sim$0.1 s} & \textbf{0.42 GB} \\
  \bottomrule
\end{tabular}

\vspace{6pt}
\begin{minipage}{\linewidth}
  \raggedright \footnotesize
  \textsuperscript{\textdagger} Time includes vanilla 2DGS \cite{Huang2DGS2024} training and a visibility precomputation step. For a pre-trained 2DGS asset, setup time is only about 2 seconds.
\end{minipage}
\end{table*}

\subsection{Relighting Efficiency Comparison}
Table~\ref{tab:timing_transplat_compact} compares the per-scene runtime of all methods. \name{} is significantly faster than all baselines. Once the 2DGS representation is trained, only a one-time $\sim$2 second precomputation step is needed per object to perform relighting into a novel environment. We provide a detailed profiling of the computational time and memory footprint for each component of \name{} in Supplementary, Table 1.

\subsection{Ablation Study}
\begin{figure*}[t!]
  \centering
  \includegraphics[width=\linewidth]{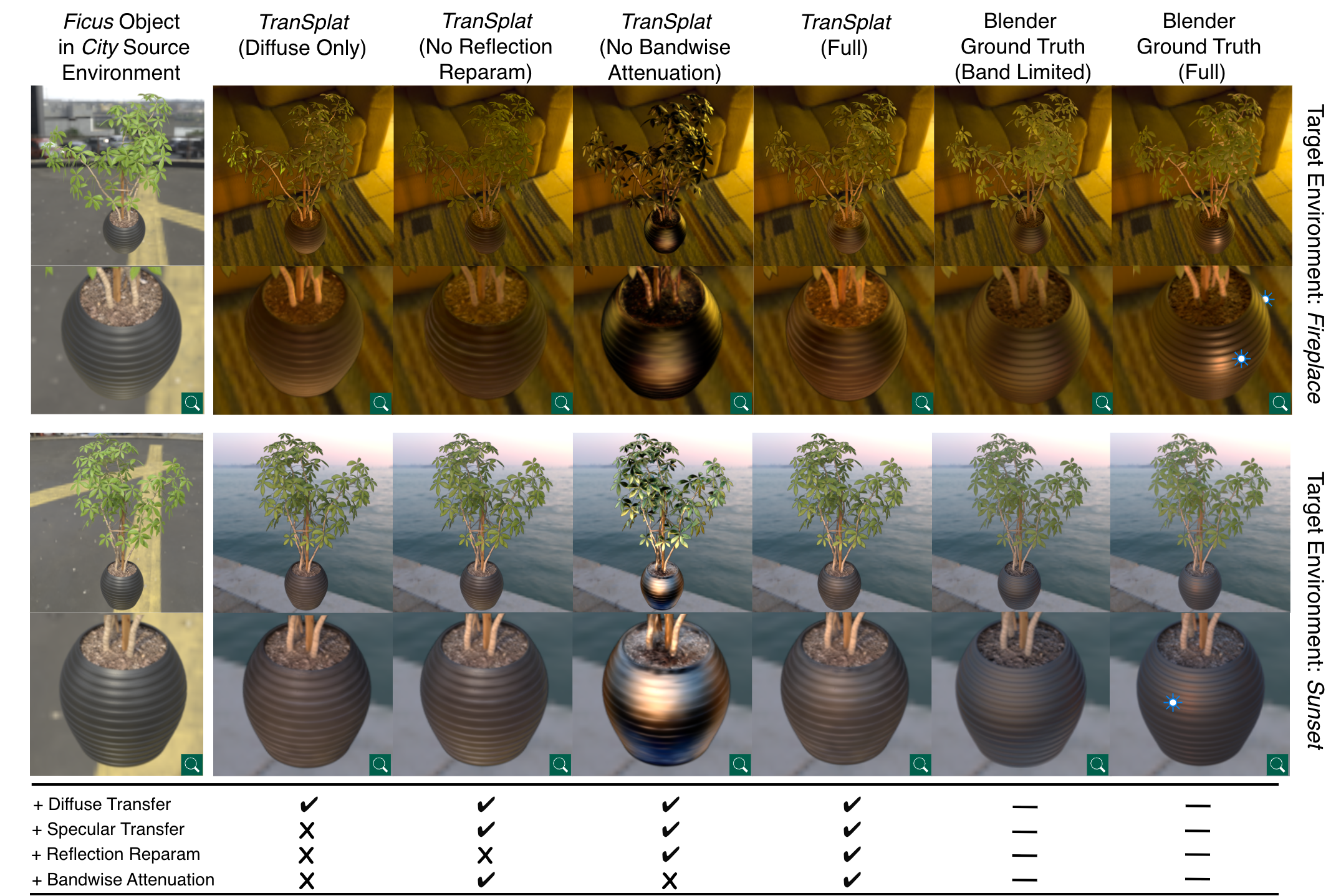}
  \caption{\textbf{Ablation demonstration of \name{}'s components on the \emph{Ficus} in the \emph{City} source environment (Column 1), from two viewpoints.} Second and fourth rows show zoomed-in versions of the first and third rows. \name{} with diffuse transfer only (\textbf{Column 2}) correctly modifies the base object color but misses specular highlights (Sec.~\ref{sec:diffuse_transfer}). \name{} without reflection reparameterization (\textbf{Column 3}) exhibits specular highlights anchored to the source-environment geometry rather than accurately following the view direction. \name{} without bandwise attenuation (\textbf{Column 4}) transfers full target-environment energy into every SH band, causing the surface to exhibit intense mirror-like artifacts. The full \name{} approach (\textbf{Column 5}) balances view-dependent highlights and correct material roughness, and closely matches the ground truth output in Blender using an HDR environment map with $l{\le}3$ SH projections (\textbf{Column 6}). The Blender ground truth using the full HDR environment map produces additional high-frequency highlights that \name{} does not capture when restricted to degree 3 SHs (\textbf{Column 7)}. The sun stickers highlight
the positions of specular highlights.}  %\textbf{\name{} (No Reflection Reparam.)} captures specularities and performs per-band BRDF attenuation. However, without reflection reparameterization, the highlights are anchored to the source-environment geometry rather than accurately following the view direction. } % \textbf{(1) Diffuse Transfer Only} correctly shifts the overall base color but reproduces no specular highlights (Sec.~\ref{sec:diffuse_transfer}). 
  %\textbf{(2) + Specular + Bandwise Attenuation} adds higher-order SH transfer and per-band BRDF attenuation. However, without reflection reparameterization, the highlights are anchored to the source-environment geometry rather than accurately following the view direction. 
  %\textbf{(3) + Specular + Reflection Reparam.} parameterizes the SH convolution in the per-Gaussian reflection frame to correct the highlight position (Sec.~\ref{sec:specular_transfer}). Because it omits the per-band BRDF attenuation $A_l^j$ (Eq.~\ref{eq:al_extraction}), it transfers full target-environment energy into every SH band, causing the surface to incorrectly appear nearly mirror-like. 
  %\textbf{(4) \name{} (Full)} applies the complete method, balancing both view-dependent highlights and correct material roughness. 
  %\textbf{(5) Blender Ground Truth (Band Limited)} renders the Blender reference using the same $l{\le}3$ SH projection of the HDR environment as incident illumination, the identical frequency ceiling imposed by our SH representation. \name{}'s output closely matches this reference. 
  %\textbf{(6) Blender Ground Truth (Full)} uses the complete HDR environment and produces sharp, high-frequency highlights that exceed what any $l{=}3$ SH method can represent, demonstrating the residual discrepancies of \name{} when restricted to degree 3 SHs.}
  \label{fig:ablation}
\end{figure*}

We next demonstrate the impact of each \name{} component with an ablation study on \emph{Ficus} (Fig.~\ref{fig:ablation}). Omitting reflection reparameterization yields specular highlights that incorrectly adhere to the source geometry, while removing band-wise attenuation causes severe, mirror-like overexposure. The full framework is necessary to achieve both view-consistent highlights and realistic material roughness. Second, we quantitatively studied the effect of maximum SH degree on (source environment) novel view synthesis (NVS) and (target environment) relighting performance in Table~\ref{tab:sh_order_ablation}. Higher SH degrees (e.g., $l{=}8$) slightly improve NVS, but provide negligible gains in relighting. However, any potential improvement in accuracy is offset by a significant (${\sim}5\times$) degradation to rendering FPS. Additional diffuse/specular decomposition results and L=3/L=8 comparisons are shown in Supplementary, Sec. 1.

\begin{table}[t!]
\centering
\small
\renewcommand{\arraystretch}{1.0}
\caption{\textbf{Ablation results on max SH appearance coefficient degree $l$.} We report mean metrics over all six objects for \emph{Fireplace}, \emph{Forest}, and \emph{Sunset} environments for both (source environment) novel view synthesis (NVS), and (target environment) relighting tasks. Increasing the SH order yields negligible gains in relighting quality with significant reduction in rendering speed (frames per second, or FPS).}
\label{tab:sh_order_ablation}
\begin{tabular*}{0.45\textwidth}{@{\extracolsep{\fill}}ll ccc@{}}
\toprule
\textbf{Task} & \textbf{Metric} & $l{=}3$ & $l{=}5$ & $l{=}8$ \\
\midrule
\multirow{3}{*}{NVS}
  & PSNR$\uparrow$  & 42.84 & 43.71 & \textbf{44.24} \\
  & SSIM$\uparrow$  & 0.99 & 0.99 & \textbf{1.00} \\
  & LPIPS$\downarrow$ & 0.01 & 0.01 & 0.01 \\
\midrule
\multirow{3}{*}{Relighting}
  & PSNR$\uparrow$  & 30.23 & 30.26 & \textbf{30.30} \\
  & SSIM$\uparrow$  & 0.95 & \textbf{0.96} & 0.95 \\
  & LPIPS$\downarrow$ & 0.05 & 0.05 & 0.05 \\
\midrule
Rendering & FPS$\uparrow$ & \textbf{101.25} & 40.54 & 21.43 \\
\bottomrule
\end{tabular*}
\end{table}

%% file: sec/5_discussion.tex
\section{Discussion and Limitations}
Our results demonstrate that \name{} produces better relighting accuracy compared to recent baseline methods. Explicit inverse-rendering pipelines can suffer from optimization instabilities that introduce color shifts or artifacts~\cite{li2025recap,R3DG2023}. Furthermore, methods specialized for reflective surfaces, like GaussianShader~\cite{jiang2024gaussianshader}, tend to over-predict specular lobes, causing diffuse objects to appear unnaturally glossy when relit. Meanwhile, generative 2D diffusion priors face resolution constraints and long inference times~\cite{jin2024neural_gaffer}. In contrast, \name{} offers a practical balance of shading fidelity, multi-view consistency, and computational efficiency, yielding significantly faster relighting after the initial scene modeling and precomputation stages.

\name{} has several limitations worth considering based on application properties. First, our analytical formulation assumes a radially symmetric BRDF and does not model non-local lighting effects such as inter-object reflections, which heavy physically-based rendering pipelines can capture at greater computational cost. Second, and most fundamentally, the SH basis acts as an inherent low-pass filter on the sphere. Faithfully reproducing near-mirror reflections or other extremely high-frequency specular effects requires an angular bandwidth that standard SH degrees simply cannot provide. As established in our ablation study, attempting to force sharper reflections by increasing the SH degree is computationally inefficient and yields diminishing returns. For this reason, we stick with the vanilla 2DGS default of $l=3$, accepting that it cannot perfectly resolve highly concentrated specular lobes. A natural extension would replace the SH representation with Spherical Gaussians (SGs), which support arbitrarily sharp specular lobes. However, fitting per-Gaussian SG lobes requires explicit BRDF decomposition, prompting a future investigation into the tradeoffs of sacrificing our BRDF-free efficiency in favor of ultra-high-frequency material accuracy. In summary, \name{} establishes a lightweight, complementary path forward for object relighting in the GS framework. 

%By prioritizing instant, perceptually realistic results over the heavy computational costs of explicit modeling, it provides a practical strategy that can be deployed in isolation for real-time applications, or combined with complex physical priors of future methods.

%% file: main.bib
@String(CVPR  = {IEEE Conf. Comput. Vis. Pattern Recog.})

@String(ECCV  = {Eur. Conf. Comput. Vis.})

@String(TOG   = {ACM Trans. Graph.})

@String(CVPR  = {CVPR})

@String(ECCV  = {ECCV})

@String(TOG   = {ACM TOG})

@article{kerbl3Dgaussians,
      author       = {Kerbl, Bernhard and Kopanas, Georgios and Leimk{\"u}hler, Thomas and Drettakis, George},
      title        = {3D Gaussian Splatting for Real-Time Radiance Field Rendering},
      journal      = {ACM Transactions on Graphics},
      number       = {4},
      volume       = {42},
      month        = {July},
      year         = {2023},
      url          = {https://repo-sam.inria.fr/fungraph/3d-gaussian-splatting/}
}

@article{barron2022mipnerf360,
  title={Mip-NeRF 360: Unbounded Anti-Aliased Neural Radiance Fields},
  author={Barron, Jonathan T. and Mildenhall, Ben and Verbin, Dor and Srinivasan, Pratul P. and Hedman, Peter},
  booktitle={CVPR},
  year={2022}
}

@inproceedings{martin2021nerf,
  title={Nerf in the wild: Neural radiance fields for unconstrained photo collections},
  author={Martin-Brualla, Ricardo and Radwan, Noha and Sajjadi, Mehdi SM and Barron, Jonathan T and Dosovitskiy, Alexey and Duckworth, Daniel},
  booktitle={Proceedings of the IEEE/CVF conference on computer vision and pattern recognition},
  pages={7210--7219},
  year={2021}
}

@techreport{debevec2004estimating,
title = {Estimating Surface Reflectance Properties of a Complex Scene under Captured Natural Illumination},
author = {Debevec, Paul and Tchou, Chris and Gardner, Andrew and Hawkins, Tim and Poullis, Charis and Stumpfel, Jessi and Jones, Andrew and Yun, Nathaniel and Einarsson, Per and Lundgren, Therese and Fajardo, Marcos and Martinez, Philippe},
institution = {University of Southern California Institute for Creative Technologies},
number = {ICT TR 06 2004},
year = {2004}
}

@book{marschner1998inverse,
  title={Inverse rendering for computer graphics},
  author={Marschner, Stephen Robert},
  year={1998},
  publisher={Cornell University}
}

@inproceedings{boivin2001image,
  title={Image-based rendering of diffuse, specular and glossy surfaces from a single image},
  author={Boivin, Samuel and Gagalowicz, Andr{\'e}},
  booktitle={Proceedings of the 28th annual conference on Computer graphics and interactive techniques},
  pages={107--116},
  year={2001}
}

@article{durand2005frequency,
  title={A frequency analysis of light transport},
  author={Durand, Fr{\'e}do and Holzschuch, Nicolas and Soler, Cyril and Chan, Eric and Sillion, Fran{\c{c}}ois X},
  journal={ACM Transactions on Graphics (TOG)},
  volume={24},
  number={3},
  pages={1115--1126},
  year={2005},
  publisher={ACM New York, NY, USA}
}

@inproceedings{ramamoorthi2001signal,
  title={A signal-processing framework for inverse rendering},
  author={Ramamoorthi, Ravi and Hanrahan, Pat},
  booktitle={Proceedings of the 28th annual conference on Computer graphics and interactive techniques},
  pages={117--128},
  year={2001}
}

@article{kajiya1986rendering,
  title={The Rendering Equation},
  author={Kajiya, James T.},
  booktitle={Proceedings of the 13th annual conference on Computer graphics and interactive techniques},
  year={1986},
  pages={4}
}

@article{R3DG2023,
    author    = {Gao, Jian and Gu, Chun and Lin, Youtian and Zhu, Hao and Cao, Xun and Zhang, Li and Yao, Yao},
    title     = {Relightable 3D Gaussian: Real-time Point Cloud Relighting with BRDF Decomposition and Ray Tracing},
    journal   = {arXiv:2311.16043},
    year      = {2023},
}

@inproceedings{yao2022neilf,
  title={Neilf: Neural incident light field for physically-based material estimation},
  author={Yao, Yao and Zhang, Jingyang and Liu, Jingbo and Qu, Yihang and Fang, Tian and McKinnon, David and Tsin, Yanghai and Quan, Long},
  booktitle={European conference on computer vision},
  pages={700--716},
  year={2022},
  organization={Springer}
}

@inproceedings{zhang2023neilf++,
  title={Neilf++: Inter-reflectable light fields for geometry and material estimation},
  author={Zhang, Jingyang and Yao, Yao and Li, Shiwei and Liu, Jingbo and Fang, Tian and McKinnon, David and Tsin, Yanghai and Quan, Long},
  booktitle={Proceedings of the IEEE/CVF International Conference on Computer Vision},
  pages={3601--3610},
  year={2023}
}

@inproceedings{Mahajan2006,
  author    = {Dhruv Mahajan and Ravi Ramamoorthi and Brian Curless},
  title     = {A Theory of Spherical Harmonic Identities for BRDF/Lighting Transfer and Image Consistency},
  booktitle = {Computer Vision -- ECCV 2006},
  editor    = {Ales Leonardis and Horst Bischof and Axel Pinz},
  series    = {Lecture Notes in Computer Science},
  volume    = {3954},
  pages     = {41--55},
  publisher = {Springer, Heidelberg},
  year      = {2006}}

@article{basri2003lambertian,
  title={Lambertian reflectance and linear subspaces},
  author={Basri, Ronen and Jacobs, David W},
  journal={IEEE transactions on pattern analysis and machine intelligence},
  volume={25},
  number={2},
  pages={218--233},
  year={2003},
  publisher={IEEE}
}

@inproceedings{Huang2DGS2024,
    title={2D Gaussian Splatting for Geometrically Accurate Radiance Fields},
    author={Huang, Binbin and Yu, Zehao and Chen, Anpei and Geiger, Andreas and Gao, Shenghua},
    publisher = {Association for Computing Machinery},
    booktitle = {SIGGRAPH 2024 Conference Papers},
    year      = {2024},
    doi       = {10.1145/3641519.3657428}
}

@inproceedings{Jin2023TensoIR,
  title={TensoIR: Tensorial Inverse Rendering},
  author={Jin, Haian and Liu, Isabella and Xu, Peijia and Zhang, Xiaoshuai and Han, Songfang and Bi, Sai and Zhou, Xiaowei and Xu, Zexiang and Su, Hao},
  booktitle={Proceedings of the IEEE/CVF Conference on Computer Vision and Pattern Recognition (CVPR)},
  year={2023}
}

@article{liang2023gs,
  title={Gs-ir: 3d gaussian splatting for inverse rendering},
  author={Liang, Zhihao and Zhang, Qi and Feng, Ying and Shan, Ying and Jia, Kui},
  journal={arXiv preprint arXiv:2311.16473},
  year={2023}
}

@article{Sloannetal2002,
  author  = {Sloan, Peter-Pike and Kautz, Jan and Snyder, John},
  title   = {Precomputed radiance transfer for real-time rendering in dynamic, low-frequency lighting environments},
  journal = {ACM Transactions on Graphics (TOG)},
  year    = {2002},
  volume  = {21},
  number  = {3},
  pages   = {527--536},
  doi     = {10.1145/566654.566612}
}

@inproceedings{jiang2024gaussianshader,
  title={Gaussianshader: 3d gaussian splatting with shading functions for reflective surfaces},
  author={Jiang, Yingwenqi and Tu, Jiadong and Liu, Yuan and Gao, Xifeng and Long, Xiaoxiao and Wang, Wenping and Ma, Yuexin},
  booktitle={Proceedings of the IEEE/CVF Conference on Computer Vision and Pattern Recognition},
  pages={5322--5332},
  year={2024}
}

@inproceedings{srinivasan2021nerv,
  title={Nerv: Neural reflectance and visibility fields for relighting and view synthesis},
  author={Srinivasan, Pratul P and Deng, Boyang and Zhang, Xiuming and Tancik, Matthew and Mildenhall, Ben and Barron, Jonathan T},
  booktitle={Proceedings of the IEEE/CVF conference on computer vision and pattern recognition},
  pages={7495--7504},
  year={2021}
}

@article{zhang2021nerfactor,
  title={Nerfactor: Neural factorization of shape and reflectance under an unknown illumination},
  author={Zhang, Xiuming and Srinivasan, Pratul P and Deng, Boyang and Debevec, Paul and Freeman, William T and Barron, Jonathan T},
  journal={ACM Transactions on Graphics (ToG)},
  volume={40},
  number={6},
  pages={1--18},
  year={2021},
  publisher={ACM New York, NY, USA}
}

@inproceedings{sang2020single,
  title={Single-shot neural relighting and svbrdf estimation},
  author={Sang, Shen and Chandraker, Manmohan},
  booktitle={Computer Vision--ECCV 2020: 16th European Conference, Glasgow, UK, August 23--28, 2020, Proceedings, Part XIX 16},
  pages={85--101},
  year={2020},
  organization={Springer}
}

@inproceedings{li2020crowdsampling,
  title={Crowdsampling the plenoptic function},
  author={Li, Zhengqi and Xian, Wenqi and Davis, Abe and Snavely, Noah},
  booktitle={Computer Vision--ECCV 2020: 16th European Conference, Glasgow, UK, August 23--28, 2020, Proceedings, Part I 16},
  pages={178--196},
  year={2020},
  organization={Springer}
}

@inproceedings{yu2019inverserendernet,
  title={Inverserendernet: Learning single image inverse rendering},
  author={Yu, Ye and Smith, William AP},
  booktitle={Proceedings of the IEEE/CVF Conference on Computer Vision and Pattern Recognition},
  pages={3155--3164},
  year={2019}
}

@inproceedings{barron2021mip,
  title={Mip-nerf: A multiscale representation for anti-aliasing neural radiance fields},
  author={Barron, Jonathan T and Mildenhall, Ben and Tancik, Matthew and Hedman, Peter and Martin-Brualla, Ricardo and Srinivasan, Pratul P},
  booktitle={Proceedings of the IEEE/CVF International Conference on Computer Vision},
  pages={5855--5864},
  year={2021}
}

@article{mildenhall2021nerf,
  title={Nerf: Representing scenes as neural radiance fields for view synthesis},
  author={Mildenhall, Ben and Srinivasan, Pratul P and Tancik, Matthew and Barron, Jonathan T and Ramamoorthi, Ravi and Ng, Ren},
  journal={Communications of the ACM},
  volume={65},
  number={1},
  pages={99--106},
  year={2021},
  publisher={ACM New York, NY, USA}
}

@inproceedings{saito2024relightable,
  title={Relightable gaussian codec avatars},
  author={Saito, Shunsuke and Schwartz, Gabriel and Simon, Tomas and Li, Junxuan and Nam, Giljoo},
  booktitle={Proceedings of the IEEE/CVF conference on computer vision and pattern recognition},
  pages={130--141},
  year={2024}
}

@inproceedings{li2025recap,
  title={ReCap: Better Gaussian Relighting with Cross-Environment Captures},
  author={Jingzhi Li and Zongwei Wu and Eduard Zamfir and Radu Timofte},
  booktitle={CVPR},
  year={2025},
}

@inproceedings{jin2024neural_gaffer,
  title     = {Neural Gaffer: Relighting Any Object via Diffusion},
  author    = {Haian Jin and Yuan Li and Fujun Luan and Yuanbo Xiangli and Sai Bi and Kai Zhang and Zexiang Xu and Jin Sun and Noah Snavely},
  booktitle = {Advances in Neural Information Processing Systems},
  year      = {2024},
}

@misc{sun2024neuralpbirreconstructionshapematerial,
      title={Neural-PBIR Reconstruction of Shape, Material, and Illumination},
      author={Cheng Sun and Guangyan Cai and Zhengqin Li and Kai Yan and Cheng Zhang and Carl Marshall and Jia-Bin Huang and Shuang Zhao and Zhao Dong},
      year={2024},
      eprint={2304.13445},
      archivePrefix={arXiv},
      primaryClass={cs.CV},
      url={https://arxiv.org/abs/2304.13445},
}

@misc{zhao2024illuminerf3drelightinginverse,
      title={IllumiNeRF: 3D Relighting Without Inverse Rendering},
      author={Xiaoming Zhao and Pratul P. Srinivasan and Dor Verbin and Keunhong Park and Ricardo Martin Brualla and Philipp Henzler},
      year={2024},
      eprint={2406.06527},
      archivePrefix={arXiv},
      primaryClass={cs.CV},
      url={https://arxiv.org/abs/2406.06527},
}

@inproceedings{stojanov2021using_toys4k,
  title={Using shape to categorize: Low-shot learning with an explicit shape bias},
  author={Stojanov, Stefan and Thai, Anh and Rehg, James M},
  booktitle={Proceedings of the IEEE/CVF conference on computer vision and pattern recognition},
  pages={1798--1808},
  year={2021}
}

@article{Troccoli2006RecoveringIA,
  title={Recovering Illumination and Texture Using Ratio Images},
  author={Alejandro J. Troccoli and Peter K. Allen},
  journal={Third International Symposium on 3D Data Processing, Visualization, and Transmission (3DPVT'06)},
  year={2006},
  pages={655-662},
  url={https://api.semanticscholar.org/CorpusID:681402}
}

@misc{guo2024prtgsprecomputedradiancetransfer,
      title={PRTGS: Precomputed Radiance Transfer of Gaussian Splats for Real-Time High-Quality Relighting}, 
      author={Yijia Guo and Yuanxi Bai and Liwen Hu and Ziyi Guo and Mianzhi Liu and Yu Cai and Tiejun Huang and Lei Ma},
      year={2024},
      eprint={2408.03538},
      archivePrefix={arXiv},
      primaryClass={cs.CV},
      url={https://arxiv.org/abs/2408.03538}, 
}

@misc{worldlabs_marble_software,
  author = {World Labs},
  title = {Marble},
  howpublished = {\url{https://marble.worldlabs.ai/}},
}

@ARTICLE{908964,
  author={Shashua, A. and Riklin-Raviv, T.},
  journal={IEEE Transactions on Pattern Analysis and Machine Intelligence}, 
  title={The quotient image: class-based re-rendering and recognition with varying illuminations}, 
  year={2001},
  volume={23},
  number={2},
  pages={129-139},
  doi={10.1109/34.908964}}

@misc{kaleta2024lumigaussrelightablegaussiansplatting,
      title={LumiGauss: Relightable Gaussian Splatting in the Wild}, 
      author={Joanna Kaleta and Kacper Kania and Tomasz Trzcinski and Marek Kowalski},
      year={2024},
      eprint={2408.04474},
      archivePrefix={arXiv},
      primaryClass={cs.CV},
      url={https://arxiv.org/abs/2408.04474}, 
}

@misc{chen2022tensorftensorialradiancefields,
      title={TensoRF: Tensorial Radiance Fields}, 
      author={Anpei Chen and Zexiang Xu and Andreas Geiger and Jingyi Yu and Hao Su},
      year={2022},
      eprint={2203.09517},
      archivePrefix={arXiv},
      primaryClass={cs.CV},
      url={https://arxiv.org/abs/2203.09517}, 
}

@article{wang2004image,
  title={Image quality assessment: from error visibility to structural similarity},
  author={Wang, Zhou and Bovik, Alan C and Sheikh, Hamid R and Simoncelli, Eero P},
  journal={IEEE transactions on image processing},
  volume={13},
  number={4},
  pages={600--612},
  year={2004},
  publisher={IEEE}
}

@inproceedings{zhang2018unreasonable,
  title={The unreasonable effectiveness of deep features as a perceptual metric},
  author={Zhang, Richard and Isola, Phillip and Efros, Alexei A and Shechtman, Eli and Wang, Oliver},
  booktitle={Proceedings of the IEEE conference on computer vision and pattern recognition},
  pages={586--595},
  year={2018}
}
